\pdfoutput=1
\documentclass[11pt]{article}
\usepackage[preprint]{acl}
\usepackage{times}
\usepackage{multirow}
\usepackage{graphicx}
\usepackage{xspace}
\usepackage{booktabs} 
\usepackage{tcolorbox}
\usepackage[hypcap=false]{caption}
\usepackage{latexsym}
\usepackage[T1]{fontenc}
\usepackage[utf8]{inputenc}
\usepackage{microtype}
\usepackage{amsmath}
\usepackage{inconsolata}
\usepackage{graphicx}
\usepackage{xcolor}
\usepackage{makecell} 
\usepackage{array}
\usepackage{colortbl}
\usepackage{xcolor}
\definecolor{pos}{RGB}{215,48,39}   
\definecolor{neg}{RGB}{69,117,180}  
\definecolor{udqbg}{RGB}{219,209,233}
\newcommand{\good}[1]{\textcolor{pos}{#1}}
\newcommand{\bad}[1]{\textcolor{neg}{#1}}
\usepackage[table]{xcolor}
\usepackage{comment}

\definecolor{mygreen}{RGB}{0, 153, 0}
\newcommand{\method}{\textsc{\textit{CARRIAGE}}\xspace}

\title{Culinary Crossroads: A RAG Framework for Enhancing Diversity in Cross-Cultural Recipe Adaptation}

\author{
  Tianyi Hu\textsuperscript{1} \quad
  Andrea Morales-Garzón\textsuperscript{3} \quad
  Jingyi Zheng\textsuperscript{4} \quad
  Maria Maistro\textsuperscript{2} \quad
  Daniel Hershcovich\textsuperscript{2} \\
  \textsuperscript{1} Aarhus University \quad
  \textsuperscript{2} University of Copenhagen \\
  \textsuperscript{3}  Dept. of Computer Science and Artificial Intelligence, University of Granada \\
  \textsuperscript{4} Hong Kong University of Science and Technology (Guangzhou)  \\
  \texttt{tenney.hu@cs.au.dk} \quad \texttt{amoralesg@decsai.ugr.es} \quad
  \texttt{jzheng029@connect.hkust-gz.edu.cn} \\
  \texttt{\{mm, dh\}@di.ku.dk}
}

\begin{document}
\maketitle

\begin{abstract}
    In cross-cultural recipe adaptation, the goal is not only to ensure cultural appropriateness and retain the original dish’s essence, but also to provide diverse options for various dietary needs and preferences.
    Retrieval-Augmented Generation (RAG) is a promising approach, combining the retrieval of real recipes from the target cuisine for cultural adaptability with Large Language Models (LLMs) for relevance. 
    However, it remains unclear whether RAG can generate diverse adaptation results. Our analysis shows that RAG tends to overly rely on a limited portion of the context across generations, failing to produce diverse outputs even when provided with varied contextual inputs.
    This reveals a key limitation of RAG in creative tasks with multiple valid answers: it fails to leverage contextual diversity for generating varied responses. 
    To address this issue, we propose \textbf{\method}, A plug-and-play RAG framework for cross-cultural recipe adaptation that enhances diversity in both retrieval and context organization. To our knowledge, this is the first RAG framework that explicitly aims to generate highly diverse outputs to accommodate multiple user preferences. 
    Our experiments show that \method achieves Pareto efficiency in terms of diversity and quality of recipe adaptation compared to closed-book LLMs.

\end{abstract}

\section{Introduction}

\begin{figure}[t]
  \includegraphics[width=\linewidth]{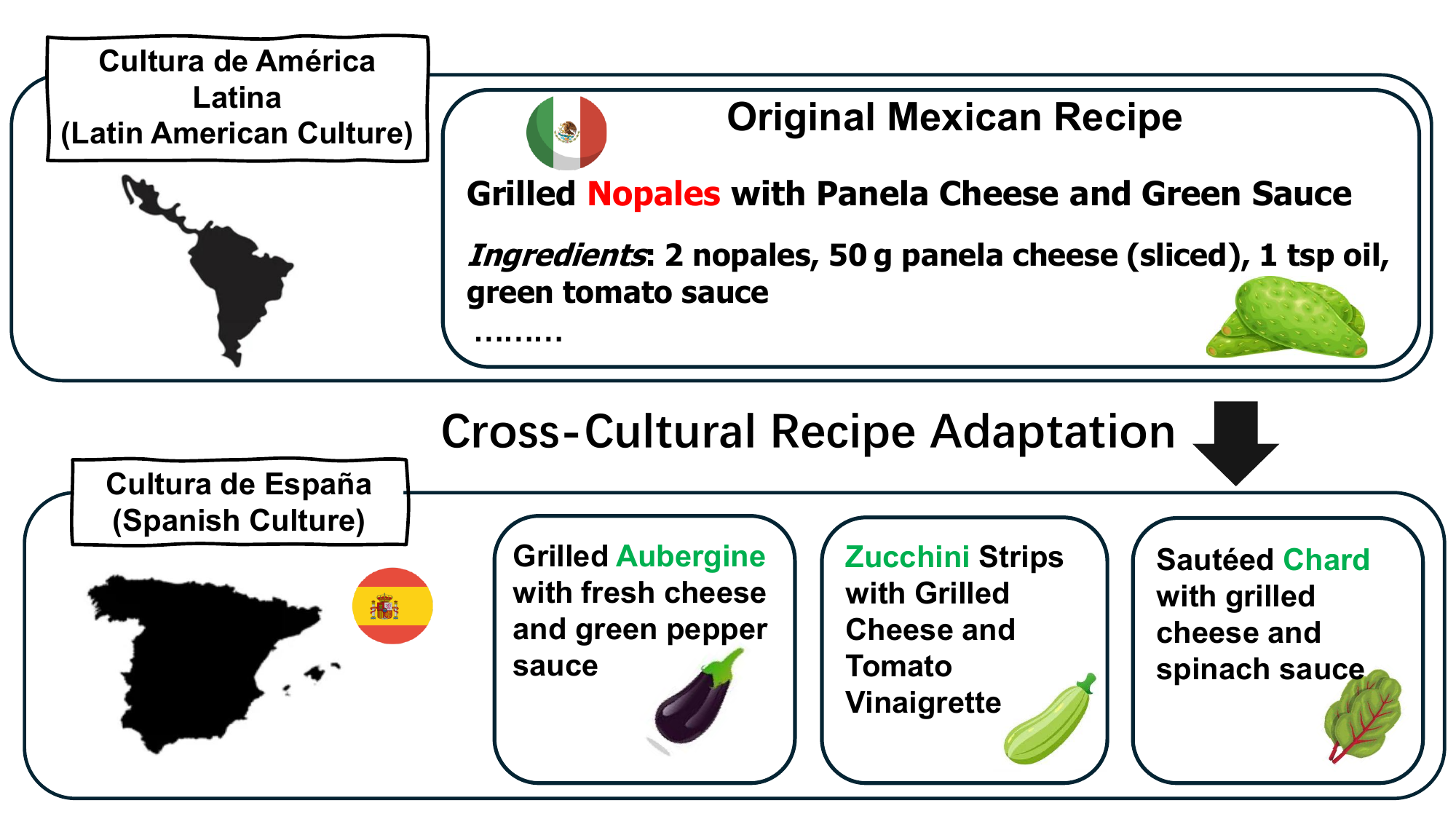} 
  \caption {A typical example of diverse choices in cross-cultural recipe adaptation. \textcolor{red}{Nopal} is a common ingredient in Mexican cuisine. However, adapting it to Spanish culinary preferences requires certain modifications, and there are many different alternatives (\textcolor{mygreen}{highlighted in green}) to consider. This illustrates the necessity for our model to account for diversity in the adaptation process.}
    \label{fig:fig1}
    \vspace{-15pt}
\end{figure}

Food serves as a powerful cultural lens, reflecting a society's history, values, and traditions \cite{hauser2023eating}. Building on this perspective, cross-cultural recipe adaptation \cite{10.1162/tacl_a_00634, hu2024bridging, pandey-etal-2025-culturally} has been proposed to bridge cultural gaps in cuisine. This task uses language models to transform an input recipe from a source culture into a version that preserves the essence of the original while aligning with the culinary norms and dietary preferences of a target culture.

While cultural alignment is crucial, food choices are also shaped by personal preferences, which can vary significantly even within the same cultural context \citep{vabo2014relationship}. At the same time, the rich diversity of recipes supports creativity in cooking \citep{mccabe2015creativity}. As shown in Figure~\ref{fig:fig1}, cross-cultural recipe adaptation often involves multiple possible substitutions. Ensuring diversity in these adaptations allows for a broader range of outputs, better serving the varied preferences of individual users. However, existing works have largely focused on generating high-quality adapted recipes \cite{10.1162/tacl_a_00634,hu2024bridging}, with much less attention given to the diversity of the outputs.

Retrieval-Augmented Generation \cite[RAG;][]{lewis2020retrieval} is a promising approach combining external knowledge retrieval with generation. In cross-cultural recipe adaptation, it can retrieve culturally appropriate and diverse recipes from the target culture \cite{hu2024bridging}, grounding the outputs in real culinary practices while preserving the essence of the original recipe. However, despite RAG’s success in knowledge-intensive tasks \cite{gao2024re}, its effectiveness remains unclear for tasks requiring both factual grounding and creative flexibility, such as cross-cultural recipe adaptation.

In this paper, we address the following research questions: \textbf{(1)} Can a standard RAG framework generate diverse outputs when given diverse and plausible contexts? 
\textbf{(2)} What automatic metrics should we use to evaluate the quality and diversity of adapted recipes, and how correlated are they with each other?  
\textbf{(3)} How can we design a RAG framework generating outputs that are both high-quality and diverse? 

Our investigation reveals a surprising finding: for the standard RAG framework, contextual diversity does not lead to more varied outputs. On the contrary, RAG shows significantly lower diversity in adapted recipes than closed-book LLMs, even when supplied with diverse contextual inputs.

To this end, we design \method: \underline{\textbf{C}}ultural-\underline{\textbf{A}}ware \underline{\textbf{R}}ecipe \underline{\textbf{R}}etr\underline{i}eval \underline{\textbf{A}}ugmented \underline{\textbf{Ge}}neration, a novel plug-and-play RAG framework aimed at addressing the diversity challenges of cross-cultural recipe adaptation. \method is a training-free framework that retrieves diverse context and organizes it effectively to generate recipes that are both high-quality and diverse. To the best of our knowledge, this is the first RAG framework designed to produce outputs that are both diverse and high-quality. The key contributions are as follows:
\begin{itemize}
    \item We are the first to explore RAG for integrating cultural references into cross-cultural recipe generation. Our findings show that simply providing diverse recipe contexts is insufficient for RAG to generate diverse outputs.
    \item We evaluate both the diversity and quality of recipe generation from multiple perspectives using a comprehensive set of automatic metrics. In particular, we propose \textit{Recipe Cultural Appropriateness Score}, a novel automatic metric for evaluating the cultural appropriateness of generated recipes.
    \item We propose \method, a novel RAG framework featuring diversity-oriented retrieval and generation components. Experimental results show that \method outperforms baseline methods in balancing diversity and quality and achieves Pareto efficiency compared to closed-book LLMs.\footnote{Our code is available at \url{https://github.com/TenneyHu/CARRIAGE}}
\end{itemize}

\section{Related Work}

\paragraph{Recipe Cross-Cultural Adaptation}

Recipe cross-cultural adaptation ~\citep{10.1162/tacl_a_00634} involves modifying recipes to suit the dietary preferences and writing styles of the target culture. This includes not just translation, but also adjusting formats, ingredients, and cooking methods to align with cultural norms. Previous studies~\citep{10.1162/tacl_a_00634,pandey-etal-2025-culturally, zhang2024cultural} often treat recipe adaptation as a cross-cultural translation task, exploring how prompt-based LLMs can be used for Chinese-English recipe adaptation.

However, LLM-based recipe adaptation still faces challenges. \citeposs{magomere2024you} show that such methods can be misleading and may reinforce regional stereotypes. \citeposs{hu2024bridging} further identifies two main challenges: First, LLMs lack culinary cultural knowledge, leading to insufficient cultural appropriateness. Second, the adapted recipes have quality issues, such as changing ingredients without adjusting the cooking steps accordingly. They propose another way to address these issues, namely through cross-cultural recipe retrieval, which sources recipes from real cooking practices within the target culture, generally offering better quality and cultural alignment. However, compared to directly using LLMs, the retrieved recipes often have low similarity to the original.

 All the above-mentioned studies primarily focus on the quality of generated results, including \textit{cultural appropriateness } and their \textit{preservation of the original}. However, they overlook the diversity of the results and do not explore the use of RAG for cross-cultural recipe adaptation. Our study emphasizes the trade-off between diversity and quality, with a particular focus on RAG-based approaches.

\begin{figure*}[t]
  \includegraphics[trim=0 90 0 160, clip, width=\linewidth]{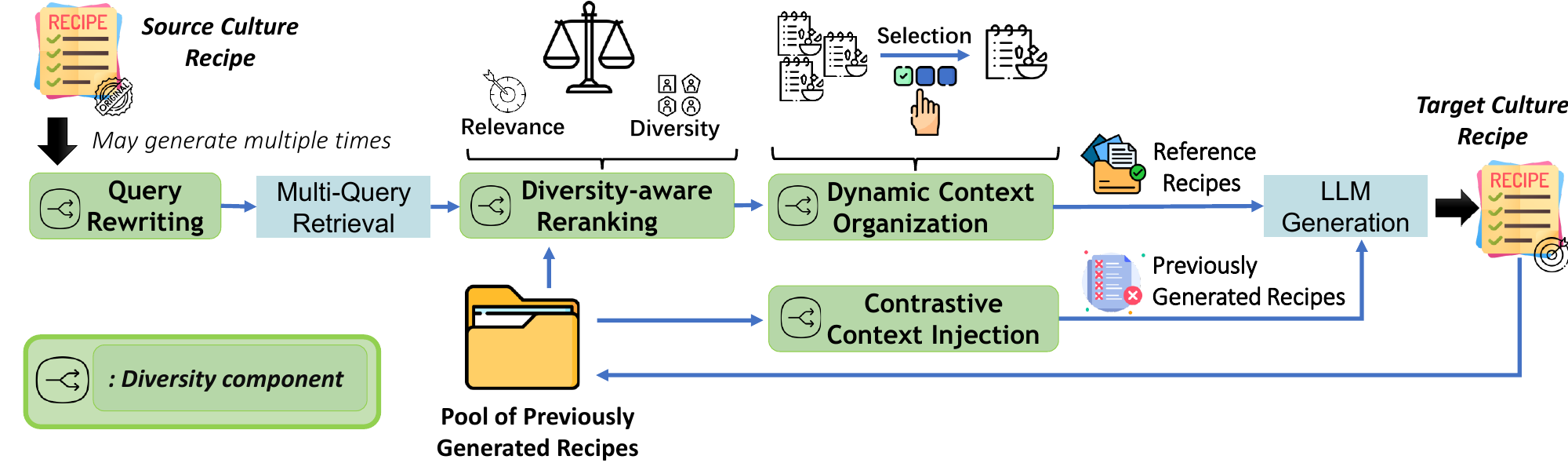} 
  \caption{Overview of \method. Diversity components are highlighted. We first enhance the diversity of retrieved results, then we enable more diverse use of contextual information via dynamic context selection, and inject contrastive context to prevent the LLM from generating outputs similar to previously generated recipes.}
  \label{fig:framework}
\end{figure*}

\paragraph{Diversity in text generation, IR, and RAG}

Previous studies~\citep{lanchantin2025diverse} have shown that post-training LLMs tend to sharpen their output probability distribution, leading to reduced response diversity. This has raised a common concern about the insufficient diversity of LLMs, particularly in creative tasks. Several stochastic sampling-based decoding methods are widely used to control the level of diversity, most notably by adjusting hyperparameters such as temperature ~\citep{shi2024thorough}. However, these methods often still fall short in achieving sufficient diversity and may lead to a rapid decline in output quality, which is another important factor to consider when measuring diversity~\citep{lanchantin2025diverse}.

In IR, retrieving text with high diversity can cover a wider range of subtopics, thereby accommodating the potentially diverse preferences of different users.
Methods such as diverse query rewriting \citep{mohankumar2021diversity} and diversity-aware re-ranking \citep{carbonell1998use,krestel2012reranking} can effectively enhance the diversity of retrieval results.
Some recent works \citep{carraro2024enhancing} have explored using LLMs to enhance diversity in re-ranking. 

In RAG, prior works have mainly focused on retrieving diverse results to obtain more comprehensive information, such as mitigating context window limitations \cite{wang2025diversity} and addressing multi-hop question answering tasks \cite{rezaei2025vendi}. These works are primarily framed as question answering, aiming to acquire comprehensive knowledge to produce a single correct answer. Consequently, the evaluation metrics emphasize answer accuracy rather than diversity. In contrast, our task naturally permits multiple valid answers. Therefore, we adopt different strategies to encourage answer diversity and use metrics that explicitly evaluate the diversity of final outputs. While prior works have largely focused on retrieving diverse contexts, our approach goes a step further by investigating how to utilize such diverse contexts to produce diverse outputs.

\section{Diversity  in Cross-Cultural Adaptation}

A standard RAG framework for cross-cultural recipe adaptation typically uses the source recipe as a query to retrieve recipes from the target culture, and then conditions generation on the top-k retrieved results to produce an adapted version. We identify four key factors that may hinder such frameworks from generating diverse adaptations.

\paragraph{C1: Missed Adaptations Due to Cultural Differences} Culturally adapted recipes often modify key ingredients or recipe names to suit cultural norms, creating semantic gaps with the source recipe. These differences hinder retrieval based solely on the original recipe, causing culturally adapted candidates to be overlooked, ultimately reducing the diversity of retrieved results.

\paragraph{C2: Lack of Diversity Awareness in Ranking} Retrieval in RAG typically ranks candidates based solely on relevance \cite{lewis2020retrieval}, which often leads to retrieving overly similar recipes, resulting in a narrow set of contexts that fail to capture the full range of culturally appropriate variations.

\paragraph{C3: Limited Contextual Variation} When users reissue the same query to seek diverse perspectives, IR models return identical results, reducing contextual diversity. Moreover, LLMs often fail to fully utilize the retrieved context \cite{liu2023lost}; they tend to focus on the same segments while ignoring others, further limiting the diversity of outputs despite the availability of varied information.

\paragraph{C4: Lack of Diversity Awareness in Generation}
Standard RAG frameworks process each query independently, failing to account for the sequential nature of user interactions. Moreover, the retrieved context provides no clear guidance for the LLM to promote output diversity.

\section{Method}
\label{sec: method_design}
To address the above challenges, we propose \method: \underline{\textbf{C}}ultural-\underline{\textbf{A}}ware \underline{\textbf{R}}ecipe \underline{\textbf{R}}etr\underline{\textbf{i}}eval \underline{\textbf{A}}ugmented \underline{\textbf{Ge}}neration. As shown in Figure~\ref{fig:framework}, we design a sequential framework that introduces diversity-enhancing components at multiple stages of retrieval and generation to improve overall diversity. First, we apply \textit{query rewriting} to the source recipe. Then, we perform \textit{diversity-aware re-ranking} to promote diversity in the retrieved results. In the generation stage, we employ \textit{dynamic context organization} to introduce varied contextual inputs across sequential queries. Finally, we apply \textit{contrastive context injection} to provide explicit signals that enhance the output diversity of the LLM.

\paragraph{Query Rewriting}
To address C1, we employ query rewriting to retrieve more diverse results. Following previous work~\cite{hu2024bridging}, we use two approaches: regenerating the recipe title based on the recipe content and prompting to culturally adapt recipe titles for a specific target audience. Query rewriting can better retrieve recipes that are similar but differ in naming due to cultural differences, thereby enriching the contextual diversity.

\paragraph{Diversity-aware Re-ranking}
To address C2, we adopt a diversity-aware re-ranking approach. We extend the classic MMR method \citep{carbonell1998use} into a ranking function that considers not only the similarity among retrieved candidates but also their similarity to past RAG outputs, in order to balance relevance and diversity. 

We define the ranking score as follows:
\vspace{-10pt}
\begin{align*}
\text{Score}(D_i) & =  \max_{D_i \in R \setminus S} \bigg[\ \lambda \cdot \text{Rel}(D_i) \\
& - (1 - \lambda) \cdot \max_{D_j \in S \cup H} \text{Sim}(D_i, D_j) \,\bigg]
\end{align*}
\vspace{-10pt}

where \( R \) is the full set of candidate documents, \( S \) is the set of already selected documents, and \( H \) is the set of recipes previously generated by the RAG model. At each iteration, the algorithm selects the next document and adds it to \( S \), repeating until the top-$k$ results are obtained.
\( \text{Rel}(D_i) \) denotes the re-ranking relevance score of document \( D_i \) to the query, \( \text{Sim}(D_i, D_j) \) measures the similarity between document \( D_i \) and \( D_j \), and \( \lambda \in [0, 1] \) is a parameter that balances relevance and diversity. We extend the diversity term in the classic MMR algorithm to consider both the selected results \( S \) and the history of RAG recommendations \( H \), by taking the maximum similarity across \(S \cup H \).

\paragraph{Dynamic Context Organization} To address C3, we introduce a simple yet effective method for better organizing the input context. Specifically, given a context containing \( k \) retrieved recipes, \( \mathcal{C} = \{D_1, D_2, \dots, D_k\} \), we apply a sliding window of size 
\( w \) to include only a subset of recipes in each generation, ensuring the context itself is diverse. By varying the context across generations, this strategy encourages the LLM to generate different outputs based on different subsets of contextual information. We define the context subset used in the \( t \)-th generation with \( t \geq 0 \) and \( (t+1)w \leq k \).
\[
\mathcal{C_\text{reference}}^{(t)} = \{ D_{tw + 1}, D_{tw + 2}, \dots, D_{(t+1)w} \}
\]

\paragraph{Contrastive Context Injection} To address C4, we introduce a contrastive context strategy, which retrieves the LLM's previous outputs generated from the same source recipe and prompts the LLM to avoid generating similar results. We believe that contrastive context helps the LLM avoid repeating previous outputs by providing explicit signals about results already generated, thereby promoting greater overall diversity in the model’s outputs.

The effectiveness of our framework is empirically validated in Section~\ref{sec: experiments}. Details about the framework can be found in Appendix~\ref{sec:imp_details}.

\section{Metrics}
\label{sec: metric}
Our evaluation metrics focus on two key aspects: \textit{diversity} and \textit{quality}. To assess diversity, we consider factors such as \textit{lexical}, \textit{semantic}, and \textit{ingredient diversity} from a per-input perspective. As a trade-off, we evaluate quality from two dimensions: the \textit{preservation} of the source recipe, and \textit{cultural appropriateness} for users in the target culture.

\subsection{Diversity}
\citeposs{kirk2023understanding} have proposed two paradigms for measuring diversity: \textit{across-input} (over pairs of one input and one output) and \textit{per-input} diversity (one input, several outputs). 
Per-input diversity helps us investigate whether a single recipe can be adapted into multiple variants to meet different dietary preferences, while across-input diversity assesses whether the generated recipes collectively exhibit a diverse range of linguistic patterns. Because our investigation primarily focuses on whether a single recipe can be adapted into diverse variations to meet a broader range of needs, we adopt the \textit{per-input } diversity setting as our main experimental focus. The \textit{across-input} diversity setting is discussed further in Section~\ref{sec:discussion}.

For a diversity metric \( \mathcal{D} \), under model configuration \( c \), \( \mathcal{A} \) denotes a set of adapted recipes, containing \( N \) source recipes, we define
\(
\mathcal{A}^i_c = \{ a^i_{c,1}, a^i_{c,2}, \ldots, a^i_{c,K} \}
\)
as the set of \( K \) adaptations for the \( i \)-th source recipe under configuration \( c \). The per-input diversity is defined as follows:
\vspace{-10pt}
\begin{equation*}
\text{PerInputDiversity}_{\mathcal{D}}(c) := \frac{1}{N} \sum_{i=1}^{N} \mathcal{D}\left(\mathcal{A}^i_c\right)
\end{equation*}
\vspace{-10pt}

\begin{table*}[ht!]
\centering
\small
\setlength{\tabcolsep}{3pt} 
\begin{tabular}{
    >{\centering\arraybackslash}p{1.1cm} |
    >{\raggedright\arraybackslash}p{3.9cm}|  
    >{\centering\arraybackslash}p{1.7cm}
    >{\centering\arraybackslash}p{1.9cm}
    >{\centering\arraybackslash}p{1.8cm}|
    >{\centering\arraybackslash}p{2.0cm}
    >{\centering\arraybackslash}p{2.0cm}
}
\toprule
\textbf{} & \multicolumn{1}{c}{\multirow{2}{*}{\centering \normalsize\textbf{Method}}} & \multicolumn{3}{|c}{\textbf{Diversity (\(\uparrow\))}} & \multicolumn{2}{|c}{\textbf{Quality (\(\uparrow\))}} \\
\cmidrule(lr){3-5} \cmidrule(lr){6-7}
& & \textbf{Lexical} & \textbf{Ingredient} & \textbf{Semantic} & \textbf{CultureScore} & \textbf{BERTScore} \\
\midrule

\multirow{3}{*}{\rotatebox[origin=c]{90}{\makecell{\textit{Closed-}\\\textit{Book}\\\textit{LLMs}}}}
& Llama3.1-8B & 0.557 & 0.667 & 0.232 & 0.451 & 0.404 \\
& Qwen2.5-7B & 0.551 & \cellcolor{neg!20}0.531 & 0.247 & \cellcolor{neg!20}0.404 & \cellcolor{red!20}0.439 \\
& Gemma2-9B & 0.538 & 0.639 & \cellcolor{neg!20}0.196 & 0.468 & \cellcolor{neg!20} 0.370 \\
\midrule

\multirow{3}{*}{\rotatebox[origin=c]{90}{\textit{IR}}}
& JINA-ES & \cellcolor{red!20}0.742 & \cellcolor{red!20}0.937 & \cellcolor{red!20}0.459 & \cellcolor{red!20}0.511 & \cellcolor{neg!20}0.295 \\
& CARROT & \cellcolor{red!20}0.735 & \cellcolor{red!20}0.925 & \cellcolor{red!20}0.462 & \cellcolor{red!20}0.512 & \cellcolor{neg!20}0.301 \\
& CARROT-MMR & \cellcolor{red!20}0.741 & \cellcolor{red!20}0.941 & \cellcolor{red!20}0.527 & \cellcolor{red!20}0.503 & \cellcolor{neg!20}0.298 \\
\midrule

\multirow{4}{*}{\rotatebox[origin=c]{90}{\textit{RAG}}}
& Vanilla-LLaMA RAG &  \cellcolor{neg!20}0.518 & \cellcolor{red!20}0.748 & \cellcolor{neg!20}0.155 & \cellcolor{neg!20}0.383 & \cellcolor{red!20}0.551 \\
& CARROT-LLaMA RAG &  \cellcolor{neg!20}0.525 & \cellcolor{red!20}0.765 & \cellcolor{neg!20}0.152 & \cellcolor{neg!20}0.385 & \cellcolor{red!20}0.545 \\
& CARROT-MMR-LLaMA RAG &  \cellcolor{neg!20}0.520 & \cellcolor{red!20}0.748 & \cellcolor{neg!20}0.164 & \cellcolor{neg!20}0.393 & \cellcolor{red!20}0.545 \\
& CARROT-MMR-Qwen RAG & 0.532 & \cellcolor{neg!20}0.536 & 0.212 & \cellcolor{neg!20}0.402 & \cellcolor{red!20}0.448 \\
\midrule

\multirow{2}{*}{\rotatebox[origin=c]{90}{\textit{Ours}}}
& \method--LLaMA & 0.577 & \cellcolor{red!20}0.739 & \cellcolor{red!20}0.269 & 0.463 & \cellcolor{red!20}0.442 \\
& \method--Qwen & \cellcolor{red!20}0.628 & 0.676 & \cellcolor{red!20}0.303 & \cellcolor{red!20} 0.590 & \cellcolor{neg!20}0.342 \\
\bottomrule
\end{tabular}
\caption{Evaluation of diversity and quality on the \texttt{RecetasDeLaAbuel@} dataset. \good{Red} indicates significant  improvements and \bad{blue} indicates significant degradations compared with the LLaMA \textit{closed-book LLMs}. It shows that our proposed \method-LLaMA outperforms all closed-book LLMs in terms of Pareto efficiency across both diversity and quality metrics (i.e., appearing only in \good{red} and not \bad{blue}). In contrast, IR-based methods struggle with preserving the source recipe, while other RAG-based approaches tend to underperform in terms of diversity and cultural appropriateness.}
\label{Table:Result}
\vspace{-10pt}
\end{table*}

\paragraph{Lexical Diversity}
Lexical diversity is a measure of the variety of vocabulary used within a set of text. High lexical diversity indicates using a broad range of unique words, which may correspond to a wider variety of ingredients, cooking methods, and flavors. We employ \textit{Unique-n} \citep{johnson1944studies} to evaluate lexical diversity, calculated as the ratio of unique $n$-grams to the total number of $n$-grams, reflecting the proportion of distinct 
$n$-grams and indicates vocabulary richness. Following prior work~\cite{guo2024benchmarking}, we report the average Unique-n across unigrams, bigrams, and trigrams.

\paragraph{Semantic Diversity}
Semantic diversity refers to the variety of meanings within a set of texts. High semantic diversity suggests a wide range of culinary ideas. We measure per-input semantic diversity using the average pairwise cosine distance between Sentence-BERT embeddings because embedding-based semantic diversity enables a more fine-grained evaluation of variation beyond surface-level vocabulary \cite{stasaski2023pragmatically}. Specifically, for a set of $K$ adapted recipes, we define the sum of their average semantic similarity and semantic diversity to be 1. In this formulation, higher semantic similarity implies lower semantic diversity. We define semantic diversity, scaled to the range $[0, 1]$, as follows: 
\begin{equation*}
\mathcal{D}_{\text{sem}}(\mathcal{A}^i_c) = \frac{1}{\binom{K}{2}} \sum_{ j < k } \frac{1-d_{\text{cos}} \left( e(a^i_{c,j}), e(a^i_{c,k}) \right)}{2}  
\end{equation*}
where \( e \) represents embeddings of the recipe.

\paragraph{Ingredient Diversity}
Ingredient diversity measures the variation in sets of ingredients across different recipes. Ingredient choice plays a crucial role in recipe diversity \cite{borghini2015recipe}. Compared to general lexical variation, ingredient changes offer a more precise signal for capturing the key factors driving diversity in recipes. 

Recipes often describe the same ingredient in varying ways, such as differences in quantity or units of measurement. To mitigate this, we introduce \textbf{Standard Ingredients}, which retain only the ingredient name by stripping away non-essential details. Since ingredient descriptions typically follow the format <\textit{quantity}> <\textit{unit}> <\textit{ingredient name}>, we extract only the <\textit{ingredient name}> to compute ingredient diversity. The detailed procedure is provided in Appendix~\ref{sec:Metrics_details}.

To avoid the influence of differing ingredient counts across recipes, we define ingredient diversity as the ratio of unique standardized ingredients to the total number of ingredients. For a set of \( K \) adapted recipes, let the set of \textit{standardized} ingredients for each recipe be \( I_1, I_2, \dots, I_K \). We define ingredient diversity as follows:
\vspace{-5pt}
\begin{equation*}
\mathcal{D}_{\text{ing}}(\mathcal{A}^i_c) = \frac{| \bigcup_{i=1}^{K} I_i |}{\sum_{i=1}^{K} |I_i|}
\end{equation*}
\vspace{-5pt}
\subsection{Quality}

We define automatic quality metrics to serve as a trade-off when evaluating recipe diversity. Further details on the training and evaluation of the CultureScore model are provided in Appendix~\ref{sec:Metrics_details}. 

\paragraph{Source Recipe Preservation}
Following prior work \cite{10.1162/tacl_a_00634, hu2024bridging}, we employ BERTScore \cite{bert-score}, a common cosine embedding-based method for measuring the similarity between source and output recipes. Previous studies have shown that BERTScore aligns well with human evaluations in terms of source recipe preservation \cite{hu2024bridging}.

\paragraph{Cultural Appropriateness} We propose a novel metric, the \textit{Recipe Cultural Appropriateness Score} (CultureScore), to assess how well the output recipes align with the target culture. Specifically, we employ a BERT-based classifier \cite{devlin2019bert, CaneteCFP2020} to predict the country of origin of a recipe using its title and list of ingredients as input. 
The CultureScore is defined as the average predicted probability assigned by the model to the target culture across all adapted recipes, with higher scores indicating better cultural alignment. Since Latin American and Spanish recipes share the same language, the model cannot rely on linguistic cues; instead, it must learn to distinguish them based on culturally relevant features such as ingredients, flavors, and writing styles.  Given that the classification model achieves an F1-score of over 90\% in distinguishing between Latin American and Spanish recipes, we consider CultureScore a reliable proxy for assessing cultural appropriateness. The details of implementation are shown in Appendix~\ref{app: culturescore}.

\section{Experiments}
In this section, we first introduce the experimental setup, then present the main experimental results, and show the correlation analysis.
\label{sec: experiments}
\subsection{Experiment Setup}

\paragraph{Task and Dataset} Our task focuses on cross-cultural recipe adaptation among Spanish-speaking countries. Despite a shared language, significant differences in food cultures make this adaptation task challenging. This requires the model not to achieve the goal through translation, but to modify the content of the recipes.
We use the \texttt{RecetasDeLaAbuel@} dataset\footnote{The Spanish recipe corpus can be accessed at \url{https://huggingface.co/datasets/somosnlp/RecetasDeLaAbuela}.}~\citep{morales2024healthy}, which is the largest Spanish-language recipe collection. It contains 20,447 entries, from which we use features including the recipe title, ingredients, preparation steps, and country of origin. 
We define the task of cross-cultural recipe adaptation as transforming recipes from seven Latin American countries, specifically Mexico, Peru, Argentina, Chile, Colombia, Venezuela, and Uruguay, into versions that align with Spanish culinary preferences. 
We randomly selected 500 source recipes from these seven countries to serve as queries, while the retrieval corpus comprises 9,381 Spanish recipes.

\paragraph{Baselines}
We consider three groups of baselines: (1) \underline{Closed-book LLMs}: We employ three of the top-performing open-source LLMs on Spanish benchmarks:\footnote{\url{https://huggingface.co/spaces/la-leaderboard/la-leaderboard}} \texttt{LLaMA3.1-8B} \citep{grattafiori2024llama}, \texttt{Gemma2-9B} \citep{team2024gemma}, and \texttt{Qwen2.5-7B} \citep{qwen2.5}. 
(2) \underline{IR Frameworks}: We include the SOTA Spanish Sentence-BERT model \cite{reimers2019sentence}, \texttt{JINA-ES} \citep{mohr2024multi}, and \texttt{CARROT} \cite{hu2024bridging}, the leading IR framework for cross-cultural recipe retrieval, and \texttt{CARROT-MMR}, a diversity-enhanced version of CARROT that integrates diversity-aware re-ranking, as described in Section~\ref{sec: method_design}.
(3) \underline{RAG Frameworks}: We leverage RAG by combining the retrieved content of selected IR models with LLMs to perform cross-cultural recipe adaptation.

\begin{figure*}[ht]
  \includegraphics[width=\linewidth]{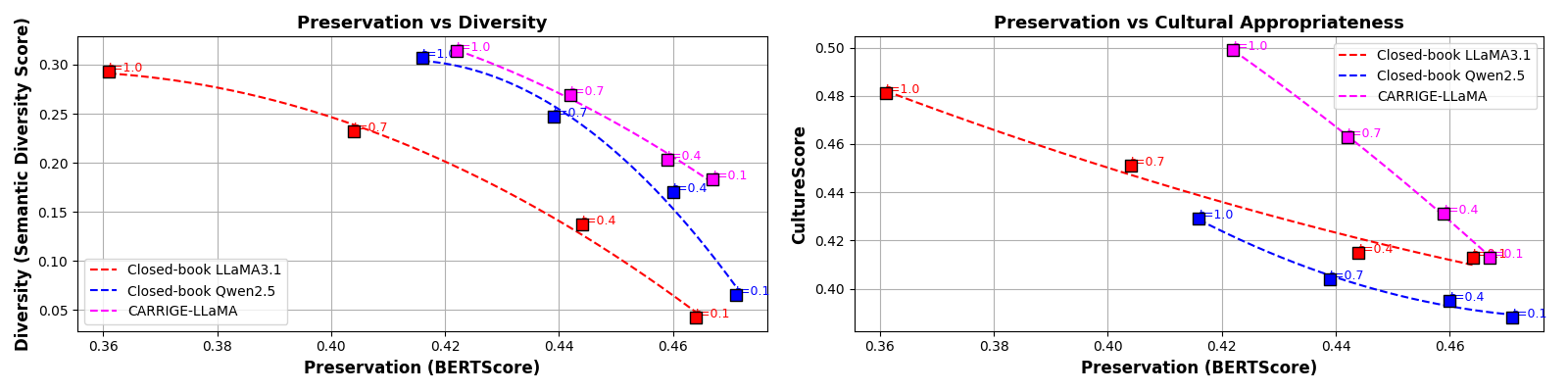} 
  \caption {Trade-offs between diversity, cultural appropriateness, and source preservation for \method and two top-performing LLMs under different temperature settings. Compared to Closed-book Qwen2.5 and LLaMA3.1 baselines, our model shows better Pareto efficiency across both key trade-offs.}
    \label{fig: trade-off}
    \vspace{-10pt}
\end{figure*}

\begin{figure*}
  \includegraphics[width=\linewidth]{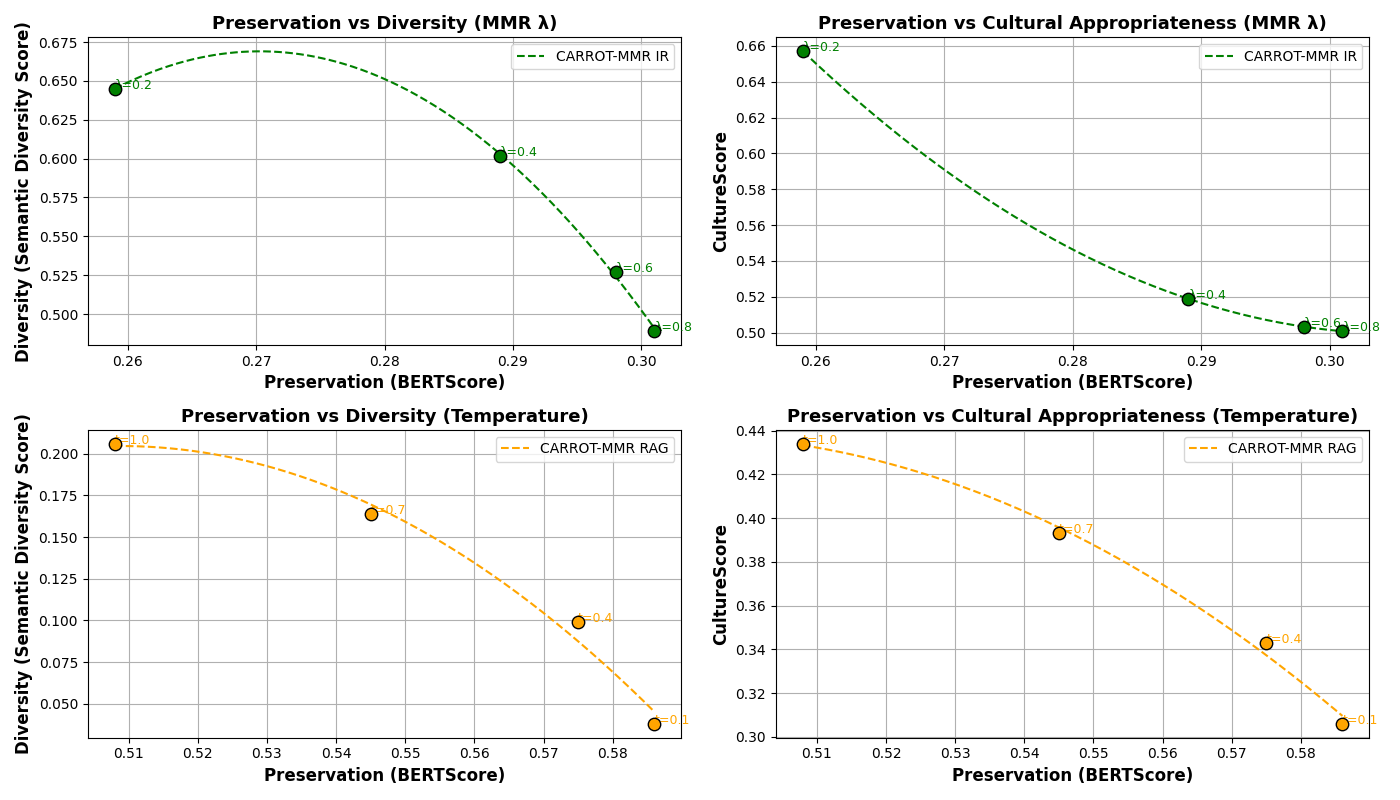} 
  \caption { Trade-off for strong baselines: CARROT-MMR-IR and CARROT-MMR-RAG. While adjusting hyperparameters allows for a trade-off, IR exhibits greatly lower source preservation, and RAG suffers from reduced diversity. Consequently, both show clear disadvantages compared to closed-book LLMs in at least one critical dimension, highlighting the strong trade-off capability of our method.}
    \label{fig: trade-off-plus}
    \vspace{-10pt}
\end{figure*}

\paragraph{Implementation Details}
    Per-input diversity is computed by generating five candidate outputs per input and evaluating the diversity within each set.
    In retrieval, we use the SOTA Spanish sentence encoder model \texttt{JINA-ES} \citep{mohr2024multi} for dense vector retrieval, and \texttt{BGE-M3} \citep{chen2024bge} as the relevance re-ranking component. Following prior work \cite{hu2024bridging}, we use the recipe title as the initial query and perform two other query rewritings to generate queries for retrieval. However, we eliminate query translation, as our scenario focuses on cross-cultural adaptation within a single language. Our main hyperparameter settings are as follows: temperature is set to 0.7, the number of retrieved contexts is 5, the diversity weight for reranking in \method is $\lambda = 0.6$, and the context sliding window size is $w = 1$. Please refer to Appendix~\ref{sec:Experiment_details} for details on other hyperparameter settings, prompts, and links to the models used.

\subsection{Main Results}

Table~\ref{Table:Result} and Figure~\ref{fig: trade-off} present our main experimental results. Appendix~\ref{sec:Supplementary} shows the result comparison across more hyperparameter configurations.

 Our study shows that both IR and LLM methods have notable limitations. Specifically, IR methods exhibit lower preservation of the source recipes, while LLMs tend to generate recipes with reduced diversity and cultural appropriateness. Among LLMs, LLaMA3.1 and Qwen2.5 demonstrate relatively strong performance, while Gemma2 shows significantly lower semantic diversity and BERTScore. Within the IR frameworks, we confirm that the diversity-aware reranking effectively enhances the semantic diversity. 
 
 We also examined the influence of hyperparameters (Table~\ref{tab:parameters} in Appendix~\ref{sec:Supplementary}). Temperature emerged as the dominant factor shaping the diversity of outputs, whereas Top-K \cite{fan2018hierarchical}, Top-P \cite{holtzman2019curious}, and Min-P \cite{nguyen2024turning} showed negligible impact on the overall performance.

 The standard RAG framework, even when provided with diverse contextual inputs and high temperature settings, yields lower semantic diversity in adapted recipes than closed-book LLMs. This highlights a key limitation of RAG in creative tasks with multiple valid outputs: it struggles to leverage contextual diversity to generate varied responses. We provide a more detailed discussion of this aspect in Section~\ref{sec:discussion}.

We evaluated our proposed framework, \method, using both LLaMA 3.1 and Qwen 2.5 as base models. Our results show that \textbf{\method-LLAMA} achieves better source recipe preservation, leading to better overall performance.
As shown in Figure~\ref{fig: trade-off}, compared to two top-performing LLMs: LLaMA 3.1 and Qwen 2.5, our model consistently achieves better \textit{Pareto efficiency} across different hyperparameter settings in the trade-offs between diversity, cultural appropriateness, and source preservation. This demonstrates that our approach achieves a more balanced and effective overall performance.

Figure~\ref{fig: trade-off-plus} illustrates the trade-offs for the strong baselines: the best IR and RAG baselines, although adjusting the hyperparameters enables trade-offs between diversity, cultural appropriateness, and source preservation, IR suffers from notably low preservation of the original recipe, while CARROT-MMR-RAG exhibits limited diversity.
As a result, both methods show distinct weaknesses when compared to closed-book LLMs and our method, further highlighting the strong trade-off capability of our method.

We also present ablation results in Appendix~\ref{app: ablation}. Removing any of the three key components: Query Rewriting, Context Organization, or Contrastive Context Steps, breaks the Pareto-efficient frontier (see Table~\ref{tab:ablation}), further confirming the effectiveness of our method.

\subsection{Correlation Study}

\begin{figure}
  \includegraphics[width=\linewidth]{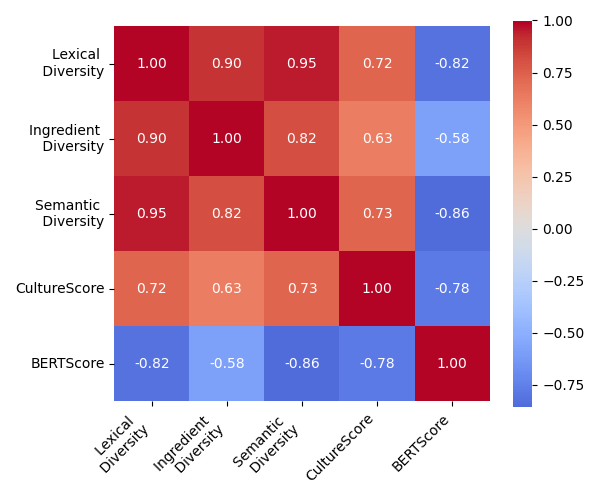} 
  \caption {Pearson correlation matrix between metrics.}
    \label{fig: correlation}
    \vspace{-10pt}
\end{figure}

\begin{figure*}[ht]
    
  \includegraphics[width=\linewidth]{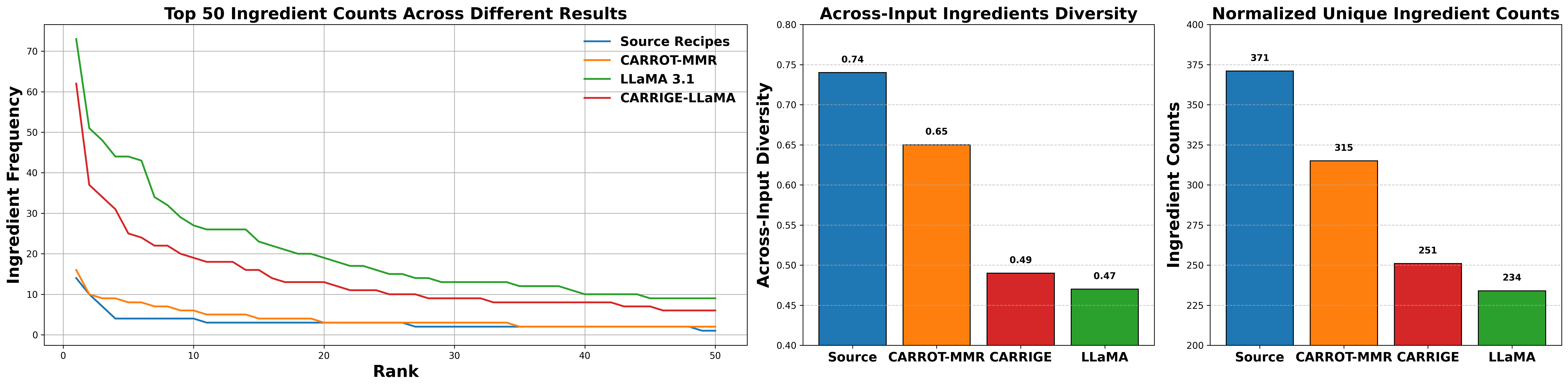} 
  \caption{Global ingredient metric across all the inputs. All adaptation methods reduce diversity compared to the original recipes, especially LLM and RAG, which notably increase the use of high-frequency ingredients.}
    \label{fig: global}
    \vspace{-10pt}
\end{figure*}

Figure~\ref{fig: correlation} illustrates the Pearson correlation coefficient between diversity and quality in model outputs. We find strong positive correlations among different aspects of recipe diversity, as well as a notable positive correlation between diversity and cultural appropriateness. In contrast, preservation of the source recipe shows negative correlations with both diversity and cultural appropriateness, since staying close to the original limits diversity and hinders cultural alignment. 

These findings highlight that preservation of the source recipe is the primary trade-off factor when aiming to generate diverse outputs. On the other hand, methods that achieve higher diversity often also lead to better cultural appropriateness, likely because IR-based methods can support both diverse recipe generation and cultural appropriateness.

\subsection{Human Alignment of CultureScore}
To assess the alignment between the BERT-based CultureScore classifier and human judgments of cultural appropriateness, we recruited two native Spanish speakers with familiarity in both Spanish and Latin American food cultures to annotate 60 samples using a 5-point scale, indicating the extent to which each recipe aligns with the cultural culinary practices of Latin American or Spanish cuisine. The quadratic-weighted Cohen's $\kappa$ between CultureScore and human annotations is 0.59, compared to 0.68 between the two annotators. Given the inherent subjectivity of cultural judgments, this level of agreement indicates a reasonable and meaningful correlation.

\section{Discussion}
\label{sec:discussion}

\paragraph{Decline in Global Ingredient Diversity} 

As shown in Figure~\ref{fig: global}, we investigate ingredient diversity in the globally aggregated adaptation results across inputs.
Specifically, we analyze the use of high-frequency ingredients, the across-input diversity, and the total number of unique ingredients.\footnote{To mitigate the impact of recipe length, we normalize by the average length of the generated ingredient lists.}

We find that all adaptation methods, even only IR, reduced global ingredient diversity, with LLM and RAG methods causing a greater reduction. 

Further analysis reveals that the reduction in diversity is largely driven by \textit{their increased reliance on high-frequency ingredients}. This results in greater similarity across adapted recipes and leads to lower diversity at the global scale. Although our work focuses on improving per-input diversity, future work could explore enhancing global across-input diversity by prompting the model to consider a broader range of ingredient substitutions beyond the most common ones.

\paragraph{Probing Diverse Contextual Utilization in RAG}
To assess whether RAG models use diverse context recipes or focus on a few, we identify the most contributing recipe for each output. The most contributing context recipe is defined as the one most semantically similar to the generated output, measured via cosine similarity between \texttt{JINA-ES} Sentence-BERT embeddings. Given a context consisting of five recipes and generating the output recipe five times, we examine whether the most relevant context recipe varies across different generations. 

As shown in Table~\ref{tab:context_count_avg}, even when provided with more diverse input contexts (e.g., \texttt{CARROT-MMR RAG}), the standard RAG framework shows only a marginal improvement in contextual diversity utilization (< \textit{7\%}). In approximately \textit{76\%} of generations, the model primarily relies on just one or two context recipes. This highlights a key limitation of the standard RAG approach in effectively leveraging the full range of retrieved information to support diverse generation. In contrast, our proposed \method, incorporating a dynamic context organization, significantly improves the model’s ability to utilize diverse context, achieving an increase of over \textit{40\%} in the contextual diversity utilization.

\begin{table}
\centering
\scriptsize
\begin{tabular}{lcccccc}
\toprule
\textbf{Method/Count} & \textbf{\#1} & \textbf{\#2} & \textbf{\#3} & \textbf{\#4} & \textbf{\#5} & \textbf{Avg.} \\
\midrule
Vanilla RAG      & 204 & 209 & 78  & 9  &  0 & 1.78 \\
CARROT RAG      & 195 & 212 & 88  & 5  &  0 & 1.81 \\
CARROT-MMR RAG  & 180 & 201 & 108 & 11 &  0 & 1.90  \\
\method RAG & 40  & 178 & 202 & 67 & 13 & \textbf{2.67} \\
\bottomrule
\end{tabular}
\caption{Distribution of \textit{most relevant context usage count} across five generations per RAG method (\#1–\#5). A higher count indicates broader contextual utilization. While increasing retrieval diversity only leads to slight improvement, our method substantially improves the diversity of context utilization, as reflected in the higher average usage count.}
\label{tab:context_count_avg}
\vspace{-15pt}
\end{table}

\section{Conclusion}

Our study provides important insights into leveraging RAG to generate diverse and high-quality recipes in the task of cross-cultural recipe adaptation, aiming to meet the varied demands of this inherently subjective and creative task. 
Our analysis reveals an important concern: the standard RAG framework, despite access to diverse contextual inputs, often fails to produce correspondingly diverse outputs. 
 To address this, we proposed a novel plug-and-play RAG framework, \method, achieving a better trade-off between output diversity and quality. Our work paves the way for future research in culturally aware and creativity-driven text generation, especially in domains that blend factual and cultural grounding with subjective variation.

\section*{Limitations}
In this work, we primarily focus on recipes from different cultures within the Spanish-speaking world. Although Spanish cuisine encompasses diverse dietary needs and presents representative challenges for cross-cultural recipe adaptation, we acknowledge that it does not fully capture the vast range of culinary traditions around the world. We encourage future work to expand the scope of cultural coverage by including a broader range of regions and cultural contexts. We believe that questions such as whether and how RAG can enhance diversity in open-ended generation are not limited to the Cross-Cultural Recipe Adaptation task; our work encourages further exploration in a broader range of open-ended settings.

Due to resource constraints, our study is based on open-source LLMs. However, we acknowledge that many proprietary, larger-scale LLMs may exhibit stronger cultural adaptation capabilities, as they potentially encode richer cultural knowledge. We hope future work will extend this line of research by incorporating more powerful LLMs. Due to resource constraints, we do not conduct human evaluations of the quality of the generated recipes. We encourage future work to incorporate human assessments to provide more accurate evaluations of recipe quality and diversity.

\section*{Acknowledgements}
Thanks to the anonymous reviewers and action editors for their helpful feedback. The research is funded by the Pioneer Centre for AI and the Novo Nordisk Foundation (NNF24OC0099109), it was also funded by ``Consejería de Transformación Económica, Industria, Conocimiento y Universidades de la Junta de Andalucía’' through a pre-doctoral fellowship program (Grant Ref. PREDOC\_00298).
We also thank the Department of Computer Science at Aarhus University for providing Tianyi Hu's travel support.

\bibliography{custom}

\appendix

\section{Details of \method Implements}
\label{sec:imp_details}

\subsection{Query Rewriting}

The model used here is Llama3.1, with the same configuration as in the main experiments. Please refer to Figure~\ref{fig:framework} for the whole pipeline and the components' position. 

\begin{tcolorbox}[colback=gray!5!white, colframe=black!75, title=Query Rewriting Prompt1: Regenerating A Title for a recipe]
Here is a recipe without a title; please create a short Spanish title for the recipe.

\vspace{0.5em}

The recipe is \textit{[recipe]}.

\vspace{0.5em}

Please only output the recipe title. Do not use quotation marks or include any explanations or additional content.

\end{tcolorbox}

\begin{tcolorbox}[colback=gray!5!white, colframe=black!75, title=Query Rewriting Prompt2: Change a New Recipe Title Based on Cultural Context]

Please rewrite the title of this recipe to align with Spanish recipe naming conventions and dietary habits.

\vspace{0.5em}

The recipe is \textit{[title]}.

\vspace{0.5em}

 Please only output the recipe title, Do not use quotation marks or include any explanations or additional content.

\end{tcolorbox}

\subsection{Prompt Setting}
Please refer to Figure~\ref{fig:framework} for the whole pipeline and the components' position. 

\begin{tcolorbox}[colback=gray!5!white, colframe=black!75, title=Prompt of \method]
Convierte la siguiente receta en una receta española para que se adapte a la cultura española, sea coherente con el conocimiento culinario español y se alinee con el estilo de las recetas españolas y la disponibilidad de ingredientes, y asegúrate de que sea diferente de las recetas en el historial proporcionado posteriormente.

A continuación se muestran algunas recetas españolas relevantes recuperadas mediante búsqueda, que pueden ser útiles para la tarea:

\textit{[context\_str]}

Dada la receta original \textit{[{query\_str}]} , utiliza las recetas anteriores recuperadas para adaptarla a una receta española.

A continuación se presentan algunos historiales; se debe EVITAR recomendar recetas similares a estas.

\textit{[{history\_str}]}

Instrucciones:

Busca recetas relevantes entre aquellas marcadas con la etiqueta [reference] para usarlas como referencia. Evita seleccionar recetas que sean similares a las marcadas con la etiqueta [history].

La receta resultante debe estar completa, incluyendo ingredientes detallados e instrucciones paso a paso. Puedes guiarte por el estilo de las recetas españolas recuperadas.

Da formato a tu respuesta exactamente de la siguiente manera:

Nombre: [Título]

Ingredientes: [Ingrediente 1] [Ingrediente 2]

Pasos:

1.

2.

...

Por favor, empieza con "Nombre: " y no añadas ningún otro texto fuera de este formato.

Mejor respuesta:

\end{tcolorbox}
\section{Details of Metrics}
\label{sec:Metrics_details}

\subsection{The Procedure of Standardized Ingredients}
To compute ingredient-level diversity, we design a rule-based cleaning function \texttt{clean\_ingredients} to extract standardized ingredient names from recipe texts. The processing steps are as follows:

\begin{itemize}
\item \textbf{Normalization:} We lowercase the input, replace fraction symbols (e.g., ½'' to 1/2''), and remove irrelevant expressions (e.g., al gusto'', para servir'') and known typos (e.g., azãºcar'' to azúcar'').

\item \textbf{Input Handling:}
\begin{itemize}
\item For \textit{AI-generated} inputs, ingredients are split using delimiters (e.g., -'', *''), and trailing parts (after commas or parentheses) are removed.
\item For \textit{human-written} inputs, we check if the ingredient field is a stringified Python list. If so, we evaluate and clean each element individually; otherwise, we treat the string as a free-form list and apply similar text cleaning.
\end{itemize}

\item \textbf{Regex-based Unit Removal:} A regular expression is applied to remove quantities, units (e.g., 200 g de''), and expressions such as 1 taza de'' or ``puñado de''.

\item \textbf{Post-processing:} We split conjunctions (e.g., ajo y cebolla'' → ajo''), remove noisy modifiers (e.g., pequeño'', mediana''), convert plural forms to singular (e.g., cebollas'' → cebolla''), and normalize characters using \texttt{unidecode}.

\item \textbf{Final Output:} The result is a cleaned list of singular, lowercase ingredient names with minimal modifiers, punctuation, or units—suitable for diversity computation.
\end{itemize}

This process is robust to variation in formatting across both model-generated and real-world recipes. The full implementation is available in the function \texttt{clean\_ingredients}.

\subsection{Implementation Details of CultureScore}
\label{app: culturescore}
To estimate the cultural appropriateness of each adapted recipe, we train a binary classifier to distinguish between Spanish and Latin American recipes. Below is a detailed explanation of the classification pipeline used to compute the proposed \textit{CultureScore}.

\begin{itemize}
\item \textbf{Model and Tokenization.} We adopt the state-of-the-art Spanish BERT model, \texttt{dccuchile/bert-base-spanish-wwm-cased}, along with its official tokenizer. This model is specifically pre-trained on large-scale Spanish corpora and has demonstrated strong performance across Spanish-language NLP tasks. For classification, we concatenate the recipe title and ingredients into a single input (e.g., ``Nombre: tortilla de patatas. Ingredientes: huevo, patata...'').

\item \textbf{Labeling and Dataset.} We use the publicly available \texttt{RecetasDeLaAbuel@} dataset, in which each recipe is annotated with its country of origin. We retain only entries with non-empty country labels and convert them into binary classification labels: Spanish (ESP) as 1 and Latin American (non-ESP) as 0. To avoid data leakage and ensure independence from our main experiments, we exclude any recipes that were used in the main generation and evaluation pipeline, training the classifier only on unused portions of the dataset.

\item \textbf{Training Setup.} The classification model is trained using HuggingFace's \texttt{Trainer} API with a 80/20 train/test split. We train for 3 epochs using a batch size of 128 and a learning rate of 2e-5.
\paragraph{CultureScore Performance.}

\item \textbf{Model Performance} We evaluate the performance of our BERT-based classifier used in computing CultureScore on a held-out test set. The model achieves an \textbf{accuracy} of 89.35\%, \textbf{F1-score} of 91.88\%, \textbf{precision} of 89.59\%, and \textbf{recall} of 94.28\%. These results indicate that the model is both precise and sensitive in distinguishing Spanish recipes from Latin American ones, confirming its reliability as a proxy for assessing cultural appropriateness.
\end{itemize}

\section{Details of Experiment}
\label{sec:Experiment_details}
\subsection{Language Models}
We used the models provided by Ollama for our LLM experiments and retrieval and reranking models from Hugging Face.

\begin{table}[ht!]
\centering
\small
\begin{tabular}{ll}
\toprule
\textbf{Model} & \textbf{Source Link} \\
\midrule
LLaMA3.1-8B & \href{https://ollama.com/library/llama3.1:8b}{\texttt{llama3.1:8b}} \\
Qwen2.5-7B & \href{https://ollama.com/library/qwen2.5:7b}{\texttt{qwen2.5:7b}} \\
Gemma2-9B & \href{https://ollama.com/library/gemma2:9b}{\texttt{gemma2:9b}} \\
Jina-V2-ES & \href{https://huggingface.co/jinaai/jina-embeddings-v2-base-es}{\texttt{jina-embeddings-v2-base-es}} \\
BGE-M3 & \href{https://huggingface.co/BAAI/bge-m3}{\texttt{bge-m3}} \\
\bottomrule
\end{tabular}
\caption{Links to all pre-trained models used in our experiments.}
\label{tab:model_links}
\end{table}
All models were used with their default configurations provided by the respective platforms, unless explicitly stated otherwise (e.g., temperature settings). All experiments were conducted on 8 NVIDIA L20 GPUs.

\subsection{Baseline Prompts Setting}

Please refer to Section~\ref{sec: experiments} for these baselines. 

\begin{tcolorbox}[colback=gray!5!white, colframe=black!75, title=Prompt of Closed-book LLMs]
Convierte la siguiente receta en una receta española para que se adapte a la cultura española, sea coherente con el conocimiento culinario español y se alinee con el estilo de las recetas españolas y la disponibilidad de ingredientes.

Dada la receta original \textit{[{query\_str}]}, utiliza las recetas anteriores recuperadas para adaptarla a una receta española.

\textbf{Instrucciones:}

La receta resultante debe estar completa, incluyendo ingredientes detallados e instrucciones paso a paso. Puedes guiarte por el estilo de las recetas españolas recuperadas.

Da formato a tu respuesta exactamente de la siguiente manera:

Nombre: [Título]

Ingredientes: [Ingrediente 1] [Ingrediente 2]

Pasos:

1.

2.

...

Por favor, empieza con "Nombre: " y no añadas ningún otro texto fuera de este formato.

\textbf{Mejor respuesta:}
\end{tcolorbox}

\begin{tcolorbox}[colback=gray!5!white, colframe=black!75, title=Prompt with Context for RAG-based LLMs]
Convierte la siguiente receta en una receta española para que se adapte a la cultura española, sea coherente con el conocimiento culinario español y se alinee con el estilo de las recetas españolas y la disponibilidad de ingredientes.

A continuación se muestran algunas recetas españolas relevantes recuperadas mediante búsqueda, que pueden ser útiles para la tarea:
 
\textit{[context\_str]}

Dada la receta original \textit{[query\_str]}, utiliza las recetas anteriores recuperadas para adaptarla a una receta española.

\textbf{Instrucciones:}  

La receta resultante debe estar completa, incluyendo ingredientes detallados e instrucciones paso a paso. Puedes guiarte por el estilo de las recetas españolas recuperadas.

Da formato a tu respuesta exactamente de la siguiente manera:

Nombre: [Título]  

Ingredientes: [Ingrediente 1] [Ingrediente 2]  

Pasos: 

1.  

2. 

...  

Por favor, empieza con \texttt{"Nombre: "} y no añadas ningún otro texto fuera de este formato.

\textbf{Mejor respuesta:}
\end{tcolorbox}
\vspace{-20pt}
\section{Supplementary experimental results}
\label{sec:Supplementary}

\subsection{Hyperparameter Study}
As shown in Table~\ref{tab:parameters}, we evaluate the impact of different decoding and retrieval parameters. Among them, temperature and the MMR balancing factor~($\lambda$) notably influence both diversity and quality, whereas parameters such as top-$k$, top-$p$, and min-$p$ exhibit relatively minor effects. 

\subsection{Ablation Study}
\label{app: ablation}
Under the same setting as in Table~\ref{Table:Result}, we added three ablation studies in Table~\ref{tab:ablation} (the removed component are in bold in the table), each corresponding to a key component of our framework. We observed that diversity, particularly the semantic diversity, shows a clear decline, especially when compared with the close-book LLM: Llama and Qwen. After ablation, the framework no longer forms a Pareto-efficient frontier. This further confirms the effectiveness of our proposed method.

\begin{table*}[ht!]
\centering
\small
\setlength{\tabcolsep}{3pt}
\begin{tabular}{
    >{\centering\arraybackslash}p{1.6cm} |
    >{\raggedright\arraybackslash}p{5.2cm}|
    >{\centering\arraybackslash}p{1.6cm}
    >{\centering\arraybackslash}p{1.7cm}
    >{\centering\arraybackslash}p{1.6cm}|
    >{\centering\arraybackslash}p{2.0cm}
    >{\centering\arraybackslash}p{2.0cm}
}
\toprule
\textbf{Setting} & \multirow{2}{*}{\centering \normalsize\textbf{Method}} & \multicolumn{3}{c|}{\textbf{Diversity (\(\uparrow\))}} & \multicolumn{2}{c}{\textbf{Quality (\(\uparrow\))}} \\
\cmidrule(lr){3-5} \cmidrule(lr){6-7}
& & \textbf{Lexical} & \textbf{Ingredient} & \textbf{Semantic} & \textbf{CultureScore} & \textbf{BERTScore} \\
\midrule

\multirow{4}{*}{\rotatebox[origin=c]{90}{Temp}}
& LLaMA3.1 (temp=0.1) & 0.347 & 0.337 & 0.042 & 0.413 & 0.464 \\
& LLaMA3.1 (temp=0.4) & 0.463 & 0.515 & 0.137 & 0.415 & 0.444 \\
& LLaMA3.1 (temp=0.7) & 0.557 & 0.667 & 0.232 & 0.451 & 0.404 \\
& LLaMA3.1 (temp=1.0) & 0.636 & 0.783 & 0.293 & 0.481 & 0.361 \\
\midrule

\multirow{4}{*}{\rotatebox[origin=c]{90}{Temp}}
& Qwen2.5 (temp=0.1) & 0.393 & 0.312 & 0.065 & 0.388 & 0.471 \\
& Qwen2.5 (temp=0.4) & 0.484 & 0.432 & 0.170 & 0.395 & 0.460 \\
& Qwen2.5 (temp=0.7) & 0.551 & 0.531 & 0.247 & 0.404 & 0.439 \\
& Qwen2.5 (temp=1.0) & 0.601 & 0.607 & 0.307 & 0.429 & 0.416 \\
\midrule

\multirow{4}{*}{\rotatebox[origin=c]{90}{Temp}}
& CARROT-MMR RAG (temp=0.1) & 0.331 & 0.345 & 0.038 & 0.306 & 0.586 \\
& CARROT-MMR RAG (temp=0.4) & 0.441 & 0.614 & 0.099 & 0.343 & 0.575 \\
& CARROT-MMR RAG (temp=0.7) & 0.520 & 0.748 & 0.164 & 0.393 & 0.545 \\
& CARROT-MMR RAG (temp=1.0) & 0.587 & 0.836 & 0.206 & 0.434 & 0.508 \\

\midrule

\multirow{4}{*}{\rotatebox[origin=c]{90}{Temp}}
& \method (temp=0.1) & 0.508 & 0.643 & 0.183 & 0.413 & 0.467 \\
& \method (temp=0.4) & 0.537 & 0.684 & 0.203 & 0.431 & 0.459 \\
& \method (temp=0.7) & 0.577 & 0.739 & 0.269 & 0.463 & 0.442 \\
& \method (temp=1.0) & 0.620 & 0.801 & 0.314 & 0.499 & 0.422 \\
\midrule

\multirow{3}{*}{\rotatebox[origin=c]{90}{Top-k}}
& LLaMA3.1 (top-k=10) & 0.557 & 0.663 & 0.227 & 0.449 & 0.405 \\
& LLaMA3.1 (top-k=40) & 0.559 & 0.664 & 0.227 & 0.450 & 0.404 \\
& LLaMA3.1 (top-k=100) & 0.557 & 0.665 & 0.228 & 0.454 & 0.405 \\
\midrule

\multirow{3}{*}{\rotatebox[origin=c]{90}{Top-p}}
& LLaMA3.1 (top-p=0.8) & 0.555 & 0.665 & 0.221 & 0.441 & 0.403 \\
& LLaMA3.1 (top-p=0.9) & 0.555 & 0.660 & 0.227 & 0.446 & 0.404 \\
& LLaMA3.1 (top-p=1.0) & 0.557 & 0.665 & 0.231 & 0.446 & 0.401 \\
\midrule

\multirow{3}{*}{\rotatebox[origin=c]{90}{min-p}}
& LLaMA3.1 (min-p=0) & 0.554 & 0.665 & 0.225 & 0.445 & 0.405 \\
& LLaMA3.1 (min-p=0.05) & 0.553 & 0.658 & 0.229 & 0.449 & 0.404 \\
& LLaMA3.1 (min-p=0.10) & 0.557 & 0.669 & 0.227 & 0.452 & 0.406 \\
\midrule

\multirow{4}{*}{\rotatebox[origin=c]{90}{MMR \(\lambda\)}}
& CARROT-MMR IR (\(\lambda=0.2\)) & 0.766 & 0.982 & 0.645 & 0.657 & 0.259 \\
& CARROT-MMR IR (\(\lambda=0.4\)) & 0.746 & 0.961 & 0.602 & 0.519 & 0.289 \\
& CARROT-MMR IR (\(\lambda=0.6\)) & 0.741 & 0.941 & 0.527 & 0.503 & 0.298 \\
& CARROT-MMR IR (\(\lambda=0.8\)) & 0.738 & 0.936 & 0.489 & 0.501 & 0.301 \\
\bottomrule

\end{tabular}
\caption{Results on different decoding and retrieval parameters. Temperature and $\lambda$ notably influence performance, while parameters such as top-$k$, top-$p$, and min-$p$ have relatively minor effects.}
\label{tab:parameters}
\vspace{-20pt}
\end{table*}

\begin{table*}[t]
\centering
\small
\begin{tabular}{lccccc}
\toprule
\textbf{Method} & \textbf{Lexical Diversity} & \textbf{Ingredient Diversity} & \textbf{Semantic Diversity} & \textbf{CultureScore} & \textbf{BERTScore} \\
\midrule
CARRIAGE-LLaMA & 0.577 & 0.739 & 0.269 & 0.463 & 0.442 \\
\quad - \textbf{Query Rewriting} & 0.575 & 0.737 & 0.244 (\textbf{-0.025}) & 0.479 & 0.445 \\
\quad - \textbf{Context organization} & 0.549 & 0.644 & 0.228 (\textbf{-0.041}) & 0.464 & 0.450 \\
\quad - \textbf{Contrastive Context Steps} & 0.583 & 0.762 & 0.231 (\textbf{-0.038}) & 0.430 & 0.453 \\
\midrule
Llama3.1-8B & 0.557 & 0.667 & 0.232 & 0.451 & 0.404 \\
Qwen2.5-7B & 0.551 & 0.531 & 0.247 & 0.404 & 0.439 \\
\bottomrule
\end{tabular}
\caption{Comparison of diversity and quality metrics across methods. Values in parentheses indicate the change relative to CARRIAGE-LLaMA.}
\label{tab:ablation}
\end{table*}

\section{Check List}
\subsection{Risk}
We identify no significant ethical or safety risks associated with our approach, as it operates on publicly available data and focuses solely on culinary adaptation. We only used publicly available datasets that do not contain personally identifiable information or offensive content, as verified by the dataset providers.

\subsection{The License For Artifacts}

All models and datasets used in this work comply with their respective open-source or research licenses. We ensure that all artifacts are used strictly within the permitted scope of their terms. The Code we released will be under a permissive open-source license, enabling reproducibility and reuse.

\subsection{Ai Assistants}
We used AI assistants (ChatGPT) solely for textual and grammatical refinement, without influencing the core content or experimental results.

\end{document}